\documentclass[lettersize,journal]{IEEEtran}
\usepackage{amsmath,amsfonts}
\usepackage{algorithmic}
\usepackage{booktabs}

\usepackage{algorithm}
\usepackage{array}
\usepackage[caption=false,font=normalsize,labelfont=sf,textfont=sf]{subfig}
\usepackage{textcomp}
\usepackage{stfloats}
\usepackage{url}
\usepackage{verbatim}
\usepackage{multirow}
\usepackage{graphicx}
\usepackage{cite}
\usepackage{hyperref}

\begin{document}

\title{G-OSR: A Comprehensive Benchmark for Graph Open-Set Recognition}

\author{Yicong Dong,
        Rundong He,
        Guangyao Chen,
        Wentao Zhang,
        Zhongyi Han,
        Jieming Shi,
        and Yilong Yin%
\thanks{Yicong Dong and Yilong Yin are with School of Software, Shandong University, Jinan, Shandong 250100, China (e-mail: yicong\_dong@mail.sdu.edu.cn; ylyin@sdu.edu.cn).}%
\thanks{Rundong He (e-mail: herundong0@gmail.com) and Jieming Shi are with The Hong Kong Polytechnic University, Hong Kong, China.}%
\thanks{Guangyao Chen is with Cornell University, Ithaca, NY, USA.}%
\thanks{Wentao Zhang is with Peking University, Beijing, China.}%
\thanks{Zhongyi Han is with King Abdullah University of Science and Technology, Thuwal, Saudi Arabia.}%
\thanks{Rundong He and Yilong Yin are the corresponding authors.}%
}



\markboth{Journal of \LaTeX\ Class Files,~Vol.~14, No.~8, August~2021}%
{Shell \MakeLowercase{\textit{et al.}}: A Sample Article Using IEEEtran.cls for IEEE Journals}


\maketitle

\begin{abstract}
Graph Neural Networks~(GNNs) have achieved significant success in machine learning, with wide applications in social networks, bioinformatics, knowledge graphs, and other fields. Most research assumes ideal closed-set environments. However, in real-world open-set environments, graph learning models face challenges in robustness and reliability due to unseen classes. This highlights the need for Graph Open-Set Recognition~(GOSR) methods to address these issues and ensure effective GNN application in practical scenarios. Research in GOSR is in its early stages, with a lack of a comprehensive benchmark spanning diverse tasks and datasets to evaluate methods. Moreover, traditional methods, Graph Out-of-Distribution Detection (GOODD), GOSR, and Graph Anomaly Detection (GAD) have mostly evolved in isolation, with little exploration of their interconnections or potential applications to GOSR. To fill these gaps, we introduce \textbf{G-OSR}, a comprehensive benchmark for evaluating GOSR methods at both the node and graph levels, using datasets from multiple domains to ensure fair and standardized comparisons of effectiveness and efficiency across traditional, GOODD, GOSR, and GAD methods. The results offer critical insights into the generalizability and limitations of current GOSR methods and provide valuable resources for advancing research in this field through systematic analysis of diverse approaches.
\end{abstract}


\section{Introduction}
\IEEEPARstart{G}{raph} learning, as a significant research direction in machine learning, has been widely applied in social network analysis, recommendation systems, bioinformatics, knowledge graphs, traffic planning, and the fields of chemistry and materials science \cite{wu2021comprehensive}. Graph Neural Networks (GNNs) have demonstrated superior performance in various node classification and graph classification tasks \cite{zhang2020methods}. These methods typically follow a closed-set setting, which assumes that all test classes are among the seen classes accessible during training \cite{bendale2016open}. However, in real-world scenarios, due to undersampling, out-of-distribution, or anomalous samples, it is highly likely to encounter samples belonging to novel unseen classes, which can significantly impact the safety and robustness of models \cite{hendrycks2016baseline}, as illustrated in \hyperlink{figpic1}{Figure 1}. As graph machine learning technologies advance across various fields, it is important to develop robust Graph Open-Set Recognition (GOSR) techniques.

\begin{figure}[!t]
\centering
\includegraphics[width=\linewidth]{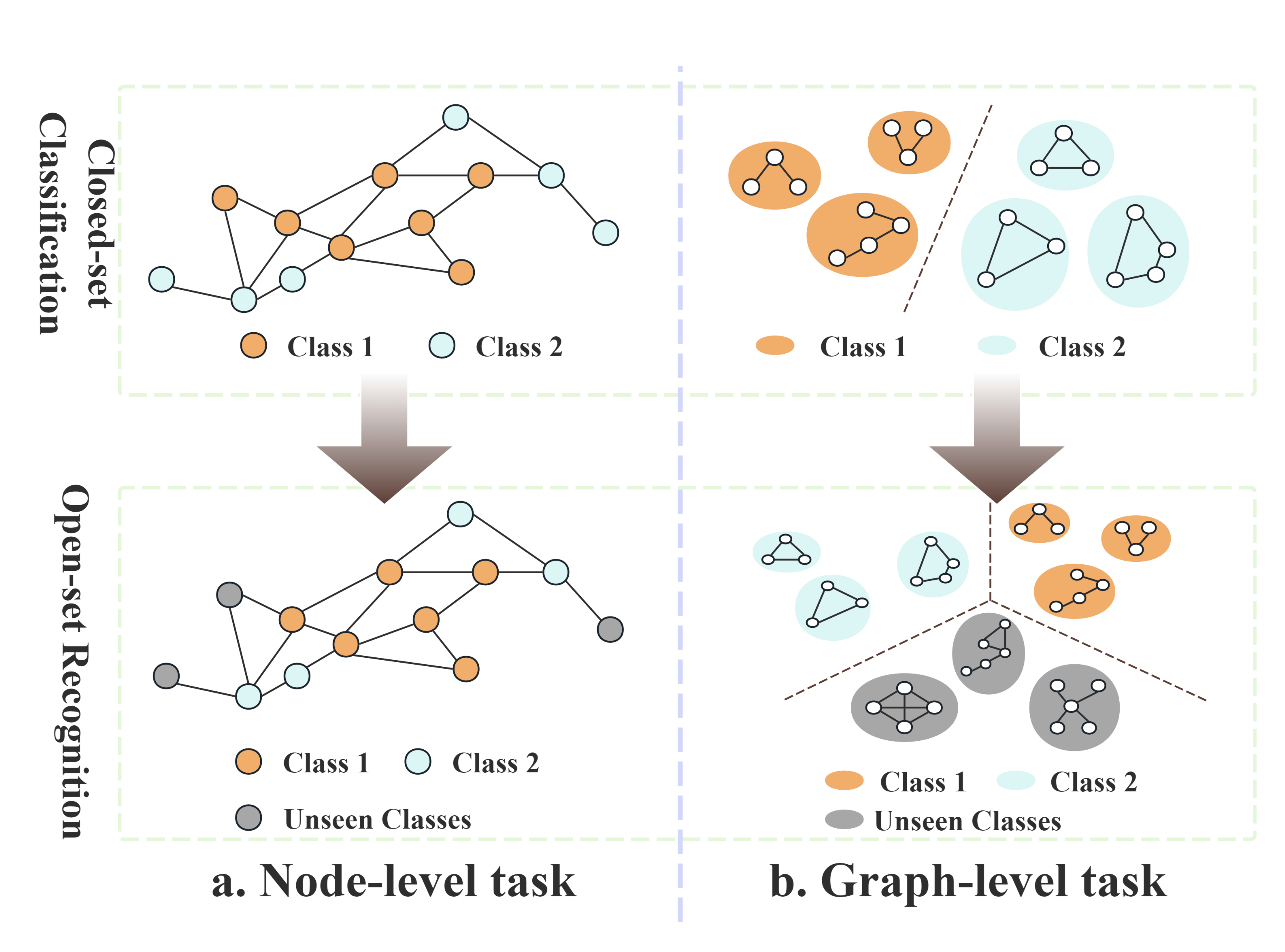}
\caption{Examples illustrating the difference between closed-set classification and open-set recognition in node-level and graph-level tasks. Closed-set classification cannot identify unseen classes, while open-set recognition can identify unseen classes and classify nodes belonging to seen classes.}
\hypertarget{figpic1}{}
\label{fig_1}
\end{figure}

To tackle the challenges that unseen classes in open-set environments pose to intelligent models, Open-Set Recognition (OSR) methods have been developed and have garnered significant research attention \cite{bendale2016towards}. However, when dealing with open-set recognition tasks for graph data, traditional OSR methods often perform poorly due to the inherent structural dependencies, heterogeneity, and complex topologies of graph data \cite{ma2021graph}. These unique characteristics require that GOSR methods understand the relationships between nodes, manage data diversity and variability, and efficiently handle large-scale and complex network topologies \cite{zhang2020methods}. Therefore, it's essential to develop effective OSR methods that leverage the unique properties of graph data, ensuring robust performance in open-set environments.

Research in GOSR mainly focuses on two tasks: node-level open-set recognition and graph-level open-set recognition, each defined by the scale of tasks. Node-level open-set recognition focuses on identifying node categories in a graph that were not encountered during training, emphasizing the local properties of nodes and their direct connections \cite{perozzi2014deepwalk, kipf2016semi}. This approach is suitable for scenarios where the characteristics and interactions of individual nodes are crucial, such as user node classification in social networks or product recommendation on e-commerce websites \cite{yang2016revisiting}. Conversely, graph-level open-set recognition aims to determine whether an entire graph belongs to an unseen category by examining the global properties of the graph, including its overall topology, node relationships, and holistic features \cite{zhang2020deep}. Common applications of this approach include the recognition of chemical molecular structures and the prediction of protein functions \cite{fout2017protein}.

One of the major challenges in GOSR lies in the absence of a fair and comprehensive benchmark for evaluating existing methods. This issue manifests in four ways:
\begin{itemize}
    \item \textbf{Limited Evaluation Across Task Levels.} Current evaluation efforts do not cover multiple task levels, which limits the comprehensiveness of assessments and fails to address the diverse application needs within graph-specific learning; 
    \item \textbf{Insufficient Cross-Domain Validation.} Inconsistencies across studies stem from the diverse range of datasets used to validate GOSR methods, which span multiple domains such as bioinformatics and social networks, each with significant differences in graph size, node types, and topological structures~\cite{dwivedi2020benchmarking}. Validating methods across datasets from multiple domains is essential to ensure their reliability and transferability;
    \item \textbf{Disconnection Between Related Fields.} Closely related areas such as Graph Out-Of-Distribution detection (GOODD) and Graph Anomaly Detection (GAD) have evolved into relatively independent research directions~\cite{ma2021comprehensive}. However, the ``unseen'' data in GOSR often appears as anomalies or OOD samples, making interaction between these fields essential~\cite{hu2020open}. Methods from GOODD and GAD may be effectively adapted to GOSR, thus facilitating innovation to address GOSR challenges. Furthermore, given the limited number of GOSR methods, comparing their performance across datasets helps establish baselines for future work.;
    \item \textbf{Variations in Technical Configurations.} Variations in technical details—such as dataset splits, unseen category settings, and hyperparameter configurations—complicate fair comparisons.
\end{itemize}

In this paper, we present a benchmarking study named G-OSR, specifically designed to evaluate GOSR methods. We utilize a diverse set of real-world datasets, including citation networks, social media, shopping websites, and molecules, to establish multiple benchmarks that cover both graph-level and node-level open-set recognition tasks. Under standardized experimental settings, we conducted extensive evaluations on various methods, including GOODD, GOSR, and GAD approaches. Additionally, we examine how the performance and robustness of different methods vary as the number of seen and unseen classes changes within the dataset's label space. By analyzing these variations, we provide valuable insights into the adaptability of these methods when faced with increasing complexity in the class distribution. Our in-depth analysis of the experimental results aims to provide valuable guidance for selecting the most effective GOSR methods and to inspire and support future research in this field, fostering the development of robust and reliable graph learning models.

Our contributions can be summarized as follows:
\begin{itemize}
\item \textbf{Comprehensive Benchmarking Study.} We introduce G-OSR, a comprehensive benchmark designed to evaluate GOSR methods at both node-level and graph-level tasks. By utilizing real-world datasets from various domains—including citation networks, social media, e-commerce platforms, and molecular graphs—this study provides a robust and comprehensive evaluation.
\item \textbf{Comprehensive Comparison Across Different Methods.} We reproduced and evaluated a diverse set of GOSR-related methods—including traditional OOD \& OSR methods, GOSR methods, GOODD methods, and GAD methods—under standardized conditions to ensure consistent and robust performance comparisons across a comprehensive benchmark.
\item \textbf{In-Depth Analysis And Insights.} Through extensive experiments and rigorous analysis, our findings offer valuable guidance for future research, providing a foundation to drive progress in the GOSR field by highlighting the strengths and limitations of current methods.
\item \textbf{Performance Analysis Across Class Settings.} Additionally, we explore the performance dynamics and robustness of various methods as the number of seen and unseen classes within the dataset label space changes, providing critical insights into the scalability and adaptability of these approaches.
\end{itemize}

\section{Related Work}\label{related_work}\subsection{Overview of Related Work}

This section succinctly revisits crucial methodologies within the Graph Open-Set Recognition (GOSR) domain. We first delve into traditional Out-of-Distribution (OOD) and Open Set Recognition (OSR) techniques, foundational for navigating unknown data in graphs. Subsequently, we explore developments in Graph OOD Detection tailored to graph structures and Graph Anomaly Detection (GAD), focusing on identifying anomalies within graph data. Each segment briefly outlines the principal theories and notable contributions, establishing a framework for our further discussions on GOSR and its diverse applications.

\subsection{Traditional Out-of-Distribution Detection \& Open-Set Recognition}

OOD detection methods and OSR methods both address the challenge of handling unknown data. While they differ in their objectives and application scenarios, they share many similarities and overlaps in their technical implementation. The primary goal of OOD detection is to identify anomalous samples that were not encountered during training and deviate from the known distribution. OSR methods extend this concept by not only accurately classifying known categories but also recognizing and rejecting samples from unseen categories. Both OOD detection and OSR rely on similar techniques, such as evaluating model uncertainty and measuring distances in feature space.

OOD detection methods can be broadly categorized into two types: classification-based methods \cite{DBLP:conf/iclr/HendrycksG17, DBLP:conf/nips/LiuWOL20, DBLP:conf/icml/WeiXCF0L22} and density-based methods \cite{ren2019likelihood, zhou2021amortized}. Classification-based OOD detection methods involve modeling the conditional distribution of in-distribution training data and designing a scoring function to measure the uncertainty of test data. Density-based methods, on the other hand, model the distribution of in-distribution data using probabilistic models and treat test data in low-density regions as OOD samples. A core challenge in OOD detection is the overconfidence in predictions for OOD samples.

To mitigate this issue, existing OOD detection methods primarily fall into two categories. The first category modifies the training process of the model to reduce overconfidence in OOD samples. For example, Wei et al. \cite{DBLP:conf/icml/WeiXCF0L22} introduced LogitNorm into the cross-entropy loss, decoupling the influence of logit norm from the training process. The second category involves introducing auxiliary OOD data or synthesizing virtual OOD data as an OOD supervision signal to fine-tune the model, thereby reducing overconfidence in OOD samples. For instance, Hendrycks et al. \cite{DBLP:conf/iclr/HendrycksMD19} used a large-scale auxiliary OOD dataset during training to help the model learn the distinction between in-distribution and OOD data, which helps the model output lower confidence scores on OOD data. Liu et al. \cite{DBLP:conf/nips/LiuWOL20} proposed an energy-based objective, which explicitly creates an energy gap by assigning lower energy to in-distribution data and higher energy to OOD data.

In many practical applications, large-scale auxiliary OOD datasets are often unavailable, and fine-tuning using such datasets is inefficient \cite{DBLP:conf/iclr/DuWCL22}. Additionally, collecting sufficiently clean large-scale auxiliary OOD data is challenging, as these datasets may contain in-distribution samples. Unlike using large-scale auxiliary OOD datasets, synthesizing virtual OOD data from in-distribution training data can significantly alleviate resource consumption issues. So far, several attempts have been made to utilize synthesized virtual OOD data to aid OOD detection \cite{DBLP:conf/nips/TackMJS20, DBLP:conf/iccv/TangMPWSGTW21}. Tack et al. \cite{DBLP:conf/nips/TackMJS20} generated virtual OOD data by applying strong data augmentation to in-distribution training data. Du et al. \cite{DBLP:conf/iclr/DuWCL22} modeled the hidden layer features of in-distribution training data as Gaussian distributions and sampled from the low-probability density regions of these Gaussian distributions as virtual OOD data.

Traditional OOD detection and OSR methods play a crucial role in establishing our GOSR benchmark. OOD and OSR techniques provide foundational methods for identifying and handling unknown data, which are directly relevant to the challenges faced by GOSR. By building on these existing methods, our work extends their principles to the graph domain.

\subsection{Graph Open-Set Recognition}
GOSR aims to identify and correctly classify graph data samples that belong to known categories encountered during the model's training while also recognizing and handling samples from novel, unseen categories. This recognition process involves determining whether an entire graph structure or individual nodes within the graph belong to a known category from the training set, or if they represent a new category that the model has not previously encountered. In the context of graph data, the concept of categories is usually related to the topology of the graph, node attributes, edge attributes, or a combination of these features. For most graph machine learning tasks, such as node or graph classification tasks, this involves the label distribution associated with the nodes or graphs. GOSR methods aim to maintain high classification accuracy for known categories while effectively identifying and managing graphs or nodes from unknown categories. This dual requirement presents a significant challenge, as the model must not only be accurate but also adaptable to new and unseen data. 
Wu et al.~\cite{wu2020openwgl} proposed \textbf{OpenWGL}, a method that leverages variational graph autoencoders for open-world graph learning, generating multiple feature vectors and automatically determining thresholds to identify unknown class nodes. Yang et al.~\cite{yang2023emp} introduced the Entropy Message Passing \textbf{(EMP)} mechanism, which combines entropy propagation with graph structure information to quantify the likelihood of nodes belonging to unknown classes, and employs entropy clustering to automatically distinguish between known and unknown class nodes.

\subsection{Graph Out-of-Distribution Detection}

Before the detailed exposition on Graph Out-of-Distribution~(GOODD) detection methods, it is important to clarify that in the graph domain, the notion of GOODD primarily treats distribution shifts as OOD without strictly confining OOD to category shifts. This perspective differs from traditional OOD detection, which predominantly focuses on category changes. Furthermore, within the field of graph analysis, the primary goals of GOODD and GOSR significantly differ. Specifically, GOSR aims to identify novel categories, emphasizing its distinct focus.

Recent advancements in GOODD have introduced novel methods tailored for graph data. \textbf{GraphDE}~\cite{li2022graphde} performs OOD detection by modeling the generative process of graph data, introducing a variational recognition model to infer the environment variable, and employing two-component mixed generative models. \textbf{GOOD-D}~\cite{liu2023goodd} utilizes a graph contrastive learning framework to detect OOD samples in graphs through hierarchical contrastive learning and disturbance-free graph data augmentation, without the need for ground truth labels. \textbf{SGOOD}~\cite{ding2023sgood} leverages subgraph structures explicitly to learn graph representations, aiding in graph-level OOD classification. \textbf{AAGOD}~\cite{guo2023aagod} enhances the input of graph neural networks with a parametric amplification matrix to distinguish graph-level OOD data. \textbf{GOODAT}~\cite{wang2024goodat} captures distinct patterns between OOD and in-distribution samples by learning informative subgraphs in the test samples through a graph masker.For node-level OOD detection, \textbf{GKDE}~\cite{zhao2020gkde} identifies OOD nodes by predicting the Dirichlet distribution of nodes. \textbf{GPN}~\cite{stadler2021gpn} employs a graph posterior network approach for node-level OOD detection. \textbf{GNNSafe++}~\cite{wu2023gnnsafe} enhances node OOD estimation through energy-based trust propagation.

\subsection{Graph Anomaly Detection}
GAD focuses on identifying anomalous elements within a graph, which could be nodes, edges, or subgraphs, whose attributes or structure significantly differ from other elements in the graph. Current anomaly detection settings often restrict the in-distribution to a single class. There is considerable overlap between graph anomaly detection and graph OOD detection, as both techniques aim to identify elements that deviate from normal patterns, aligning with the goal of OOD detection in identifying samples that do not conform to known distributions. Thus, graph anomaly detection techniques, especially those capable of in-depth analysis of graph structure and node features, may also be applicable to graph OOD detection problems.

In node-level anomaly detection, the focus is on identifying abnormal nodes within a graph based on their features and structural properties. \textbf{AnomalyDAE}~\cite{ding2019deep} integrates structural and attribute autoencoders to learn the interactions between network structure and node attributes, identifying anomalies through reconstruction error. \textbf{GAAN}~\cite{chen2020generative} utilizes generative adversarial networks to distinguish between real data and generated fake nodes for anomaly detection. \textbf{OCGNN}~\cite{wang2021one} maps nodes to the interior of a hypersphere centered on a central vector to identify anomalous nodes. \textbf{CoLA}~\cite{liu2021anomaly} models the relationship between each node and its neighboring substructures, using a graph neural network-based contrastive learning model and calculates anomaly scores through reconstruction error. \textbf{GUIDE}~\cite{zhu2023guide} learns the differences between nodes and their higher-order structures through graph attention layers, employing reconstruction error for node-level anomaly detection. \textbf{GAD-NR}~\cite{zhang2021graph} employs a neighborhood reconstruction-based graph autoencoder to identify and differentiate anomalous nodes. Finally, \textbf{CONAD}~\cite{luo2022conad} achieves node-level anomaly detection in attributed networks using siamese graph neural networks and contrastive loss. These methods demonstrate the use of advanced machine learning techniques, such as autoencoders, adversarial networks, and contrastive learning, to effectively detect anomalies at the node level.

For graph-level anomaly detection, the objective shifts to identifying entire graphs that exhibit anomalous patterns. \textbf{OCGIN}~\cite{zhang2019ocgin} represents early work in this area, utilizing one-class classification and graph neural network techniques for anomaly detection. \textbf{OCGTL}~\cite{zhang2020ocgtl} enhances performance by leveraging concepts from self-supervised learning and transformation learning. \textbf{GlocalKD}~\cite{zhang2023glocalkd} performs graph-level anomaly detection by introducing a global and local knowledge distillation framework.

These GAD methods are inherently designed for unsupervised settings, focusing on detecting anomalies without classifying within the normal class itself. However, by disregarding their ability to classify within the visible classes, we can evaluate their effectiveness in identifying unseen classes in GOSR tasks, thus assessing their potential for open-set recognition problems.

\section{Problem Definition}\label{problem_definition}\paragraph{Graph Open Set Recognition}
Machine learning models trained in a closed-world setting may incorrectly classify test samples from unknown graph classes as one of the known categories with high confidence. To address this issue, Graph Open Set Recognition (GOSR) is proposed. According to Scheirer et al. (2013) \cite{scheirer2013toward}, Open Set Recognition (OSR) involves distinguishing between known and unknown classes, where ``known known classes'' are those present during training and ``unknown unknown classes'' are those not encountered during training. Formally, GOSR requires a classifier to simultaneously: 1) accurately classify test samples from known classes, and 2) detect test samples from unknown classes.

Let $\mathcal{C}_{\text{known}}$ be the set of known classes and $\mathcal{C}_{\text{unknown}}$ be the set of unknown classes. The classifier $h$ maps graphs $G$ to these classes or to an ``unknown'' label. The classifier can be defined as:
\begin{equation}
h(G) = 
\begin{cases} 
c, & \text{if } G \in \mathcal{C}_{\text{known}} \\
\text{unknown}, & \text{if } G \in \mathcal{C}_{\text{unknown}} 
\end{cases},
\label{eq:gosr}
\end{equation}
where $c \in \mathcal{C}_{\text{known}}$ indicates classification into a known class, and $h(G) = \text{unknown}$ indicates an unknown class. Unlike Graph Out-of-Distribution Detection (GOODD), which focuses on identifying test samples that deviate from the training distribution without necessarily being from a completely unseen class, GOSR explicitly distinguishes between known and unknown classes by assigning an ``unknown'' label to samples that do not belong to any of the known classes.

\paragraph{Graph Out-of-Distribution Detection}
GOODD aims to identify test samples that either do not belong to any of the classes present in the training data or exhibit distributional shifts compared to the training data. Hendrycks and Gimpel \cite{hendrycks2017baseline} define OOD detection as the task of detecting samples that are outside the distribution of the training data. In the context of graph-specific machine learning, this typically refers to detecting graphs whose structural or attribute distributions differ significantly from those seen during training. Importantly, OOD detection should not compromise the in-distribution (ID) classification performance.

Let $p_{\text{train}}(G, y)$ be the joint distribution of graphs $G$ and labels $y \in \mathcal{C}_{\text{train}}$ in the training set, where $\mathcal{C}_{\text{train}}$ represents the set of training classes. The goal of OOD detection is to determine whether a test graph $G_{\text{test}}$ comes from $p_{\text{train}}$ or from a different distribution $p_{\text{test}}(G)$. Formally, an OOD detector $d$ can be defined as:
\begin{equation}
d(G) = 
\begin{cases} 
0, & \text{if } G \sim p_{\text{train}}(G) \\
1, & \text{if } G \not\sim p_{\text{train}}(G)
\end{cases},
\label{eq:ood}
\end{equation}
where $d(G) = 1$ indicates an out-of-distribution sample, and $d(G) = 0$ indicates an in-distribution sample. The key distinction between GOSR and GOODD lies in the goal: while GOSR aims to classify samples into known and unknown classes, GOODD is focused on identifying samples that do not conform to the training distribution, without necessarily labeling them as belonging to an unknown class.

\paragraph{Graph Anomaly Detection}

Graph Anomaly Detection (GAD) refers to the identification of anomalous patterns in graph data, which deviate from the norm. Anomalies in graphs can be classified into different types based on the nature of the abnormality. According to Akoglu et al. (2015) \cite{akoglu2015graph}, anomalies can be categorized into structural anomalies, which involve unusual patterns in the graph topology, and attribute anomalies, which involve unexpected attribute values associated with nodes or edges. Furthermore, graph anomaly detection methods can be divided into unsupervised and semi-supervised approaches based on the availability of labeled data.

Formally, let $G = (V, E)$ represent a graph, where $V$ is the set of nodes and $E$ is the set of edges. An anomaly detection model $f$ aims to identify subgraphs $G' = (V', E') \subseteq G$ that maximize the anomaly score $s(G')$. The model can be defined as:
\begin{equation}
f(G') = 
\begin{cases} 
0, & \text{if } s(G') < \theta \\
1, & \text{if } s(G') \geq \theta
\end{cases},
\label{eq:gad}
\end{equation}
where $f(G') = 1$ indicates an anomaly, and $f(G') = 0$ indicates normality. The anomaly score $s(G')$ often depends on measures such as deviation from expected node degree distributions or attribute consistency, and $\theta$ is a threshold that separates normal from anomalous subgraphs.

\paragraph{Definition Interpretation}

Drawing inspiration from the generalized OOD detection framework in computer vision, we integrate methods such as GOODD, GAD, and GOSR within the GOSR task framework, leveraging their inherent connections to address GOSR tasks. Within this unified task framework, we systematically evaluate the performance of these methods. Additionally, we conduct evaluations across different task levels, including both node-level and graph-level tasks, as well as across various domains using datasets from multiple fields. This comprehensive benchmark not only tests the effectiveness of each method individually but also provides insights into their relative strengths and weaknesses across diverse scenarios.

\section{Benchmark Design}\label{method}In this section, we provide a comprehensive overview of multi-task GOSR, covering datasets (Sec.~\ref{sec:datasets}), Methods (Sec.~\ref{sec:Methods}), and evaluation metrics (Sec.~\ref{sec:evaluation}).

\subsection{Datasets Setup}
\label{sec:datasets}

\begin{table}[t]
\centering
\caption{Benchmark Datasets for Node-level and Graph-level Open-Set Recognition}
\renewcommand{\arraystretch}{1.5} 
\setlength{\abovecaptionskip}{0.cm}
\scalebox{0.85}{
\begin{tabular}{c|c|c|c}
\hline
\multirow{2}{*}{\centering Task Level} & \multirow{2}{*}{\centering Domain} & \multirow{2}{*}{\centering Dataset} & \multirow{2}{*}{\centering Sample Size} \\ 
& & & \\ 
\hline
\multirow{8}{*}{\centering Node-level} 
& \multirow{4}{*}{\centering Citation Network} 
& Cora & 2,708 \\ \cline{3-4}
& & Coauthor-CS & 18,333 \\ \cline{3-4}
& & Coauthor-Physics & 34,493 \\ \cline{3-4}
& & Citeseer & 3,327 \\ \cline{2-4}
& \multirow{2}{*}{\centering Shopping Website} 
& Amazon-Photo & 7,487 \\ \cline{3-4}
& & Amazon-Computer & 13,752 \\ \cline{2-4}
& \centering Citation Network & Arxiv & 169,343 \\ 
\hline
\multirow{14}{*}{\centering Graph-level} 
& \centering Proteins & ENZYMES & 600 \\ \cline{3-4}
& & PROTEINS & 1,113 \\ \cline{2-4}
& \multirow{4}{*}{\centering Social Networks} 
& IMDB-MULTI & 1,500 \\ \cline{3-4}
& & IMDB-BINARY & 1,000 \\ \cline{3-4}
& & REDDIT-12K & 11,929 \\ \cline{3-4}
& & REDDIT-5K & 5,000 \\ \cline{2-4}
& \multirow{6}{*}{\centering Molecules} 
& BZR & 405 \\ \cline{3-4}
& & COX2 & 467 \\ \cline{3-4}
& & Tox21 & 7,831 \\ \cline{3-4}
& & SIDER & 1,427 \\ \cline{3-4}
& & BBBP & 2,039 \\ \cline{3-4}
& & BACE & 1,513 \\ 
\hline
\multicolumn{4}{c}{OSR Challenge on All Datasets: Unseen Classes} \\
\hline
\end{tabular}
}
\label{tab:benchmark_datasets}
\end{table}

\subsubsection{Node-level GOSR Datasets Setup}
To construct node-level GOSR benchmarks, we extensively reviewed and employed multiple graph node classification datasets of various sizes and complexities, spanning fields such as text classification, social network analysis, recommendation systems, and biomolecules. G-OSR supports six node-level GOSR benchmark tests, including \textbf{Cora}~\cite{mccallum2000automating}, \textbf{Coauthor-CS}~\cite{shchur2018pitfalls}, \textbf{Coauthor-Physics}~\cite{shchur2018pitfalls}, \textbf{Amazon-Photo}~\cite{mcauley2015image}, \textbf{Arxiv}~\cite{hu2020open}, and \textbf{Citeseer}~\cite{sen2008collective}.

\paragraph{Cora} 
The Cora dataset is a commonly used citation network dataset in the graph machine learning field. It contains 2,708 nodes representing scientific publications, categorized into seven classes, each representing a research area. Edges represent the citation relationships between the papers, where an edge exists between two papers if one cites the other. We partitioned the nodes of 4 categories in the Cora dataset as the seen classes, while the remaining 3 categories were used as the unseen classes, representing the open-set recognition scenario.

\paragraph{Coauthor-CS} 
Coauthor-CS is one of the datasets frequently used for node classification tasks in graph machine learning, reflecting the academic collaboration network within the field of Computer Science. In Coauthor-CS, nodes representing scientists are categorized into 15 classes, with edges indicating collaborative relationships between them, meaning if two scientists have co-authored a paper, there is an edge between them. We partitioned the nodes from 8 categories in the Coauthor-CS dataset as the seen classes, while the remaining 7 categories were used as the unseen classes.

\paragraph{Amazon-Photo} 
The Amazon-Photo dataset is a co-purchase network dataset from Amazon, focusing on the co-purchasing behavior of photography products. In this dataset, each node represents a product, and edges indicate co-purchase relationships between products. For the Amazon-Photo dataset, we selected 6 out of 8 node categories to form the seen classes, while the remaining two categories were used as the unseen classes.

\paragraph{CoraFull} 
The CoraFull dataset is an expanded version of the Cora dataset, containing 19,793 nodes categorized into 70 classes. Compared to Cora, CoraFull provides a richer set of node feature vectors and more finely divided academic domain category labels, offering a more challenging and realistic testing platform for node classification and open-set recognition tasks. This helps in evaluating the model's performance on larger-scale and higher granularity tasks.

\paragraph{Arxiv} 
Arxiv, or 'ogbn-arxiv', is a large-scale node classification graph dataset provided by the Open Graph Benchmark (OGB), constructed based on computer science papers from arXiv. It contains 169,343 nodes, each representing a paper in the computer science field on arXiv, with edges indicating the citation relationships between the papers. All papers are categorized into 40 classes, covering a range of areas from artificial intelligence to theoretical computer science. We used 20 categories from Arxiv as the seen classes, and the remaining 20 categories as the unseen classes.

\paragraph{Citeseer} 
The Citeseer dataset is a well-known citation network dataset frequently used in graph machine learning tasks. It comprises 3,327 nodes, each representing a scientific publication, categorized into six distinct classes corresponding to various research areas. The edges in the Citeseer dataset denote citation relationships between these publications, where an edge is present if one paper cites another. For the GOSR tasks, four of these classes can be designated as seen classes, while the remaining two classes serve as unseen classes. This setup allows for an assessment of the model’s ability to generalize and recognize new, unseen research topics based on the knowledge acquired from the seen classes.

\paragraph{Coauthor-Physics} 
The Coauthor-Physics dataset is an academic collaboration network within the field of Physics. In this dataset, nodes represent authors, and edges indicate collaborative relationships, meaning there is an edge between two nodes if the corresponding authors have co-authored a paper. The dataset categorizes nodes into five classes, each corresponding to different research areas within Physics. For the GOSR setup, three of these classes can be used as seen classes, while the remaining two categories represent unseen classes. This partitioning is ideal for evaluating the effectiveness of GOSR methods in handling the challenges posed by complex and densely connected academic collaboration networks.

\subsubsection{Graph-level GOSR Datasets Setup}

To construct graph-level GOSR detection benchmarks, we extensively reviewed and selected multiple graph classification datasets, covering various domains including social networks, biomolecules, and pharmaceutical compounds. G-OSR supports six graph-level graph GOSR benchmark tests, named after their ID datasets, including \textbf{ENZYMES} \cite{schomburg2004enzyme}, \textbf{IMDB-MULTI} \cite{yanardag2015deep}, \textbf{BBBP} \cite{martins2017bayesian}, \textbf{BZR} \cite{srinivasan1996extracting}, \textbf{Tox21} \cite{huang2016tox21}, and \textbf{REDDIT-12K} \cite{yanardag2015deep}. In each benchmark test, the OOD datasets exhibit significant semantic shifts compared to the

\paragraph{ENZYMES} 
ENZYMES is a dataset of 600 protein tertiary structures obtained from the BRENDA enzyme database. It is primarily used for enzyme classification tasks, categorized into six classes based on the type of chemical reactions catalyzed by the enzymes. In ENZYMES, each node represents an amino acid residue within the enzyme, and the edges represent interactions or connections between these amino acid residues. All classes of graphs in ENZYMES are considered seen classes. To introduce unseen classes, we utilized graphs from the PROTEIN dataset. PROTEINS is also a dataset of protein networks, where graphs are labeled as either 'Enzymes' or 'Non-enzymes.' We use graphs labeled as 'Non-enzymes' to represent unseen classes.

\paragraph{IMDB-MULTI} 
IMDB-MULTI is a relational dataset consisting of a network of 1,000 actors and actresses who have played roles in movies cataloged in IMDB. Each node represents an actor or actress, and an edge connects two nodes if they appeared together in the same movie. In IMDB-MULTI, the graphs are labeled with three genres: Comedy, Romance, and Sci-Fi. To introduce unseen classes, we utilized graphs from the IMDB-BINARY dataset. Similar to IMDB-MULTI, IMDB-BINARY is also a relational dataset, but its graphs are labeled as either 'Action' or 'Romance'. Specifically, we used the graphs labeled 'Action' from IMDB-BINARY as unseen classes, as these graphs do not belong to any category in IMDB-MULTI.

\paragraph{BBBP} 
The BBBP dataset originates from a study aimed at modeling and predicting the permeability of the blood-brain barrier (BBB). In the BBBP dataset, each node represents an atom within a molecule, and edges indicate the chemical bonds between atoms. The BBBP dataset contains binary labels that indicate whether a compound can penetrate the blood-brain barrier or not. We used graphs from the BACE dataset to introduce unseen classes. The BACE dataset is also a chemical dataset, designed for researching compounds related to Beta-Secretase 1 (BACE-1). Compared to the graphs in BBBP, the graphs in BACE represent compounds with significantly different biological activities, thus serving as unseen classes.

\paragraph{BZR} 
The BZR dataset originates from a study focused on modeling and predicting the biological activity of benzodiazepine receptor ligands. In the BZR dataset, each node represents an atom within a molecule, and edges indicate the chemical bonds between atoms. The BZR dataset contains binary labels that indicate whether a compound has a particular biological activity associated with benzodiazepine receptors. We used graphs from the COX2 dataset to represent unseen classes. The COX2 dataset is also a chemical dataset, designed for researching compounds related to Cyclooxygenase-2 (COX-2) inhibitors. Compared to the graphs in BZR, the graphs in COX2 represent compounds with significantly different biological activities, qualifying them as unseen classes.

\paragraph{Tox21} 
The Tox21 dataset is derived from the Toxicology in the 21st Century (Tox21) initiative, which aims to evaluate the toxicity of chemical compounds using high-throughput screening (HTS) techniques. Each node in the Tox21 dataset represents an atom within a molecule, with edges indicating the chemical bonds between atoms. This dataset contains binary labels that indicate whether a compound exhibits specific toxicological effects. We utilized graphs from the SIDER dataset to introduce unseen classes. The SIDER dataset, obtained from the Side Effect Resource, provides information on approved drugs and their associated adverse drug reactions. Compared to the graphs in Tox21, the graphs in SIDER represent compounds with significantly different biological activities, qualifying them as unseen classes.

\paragraph{REDDIT-12K} 
REDDIT-12K contains 11,929 graphs, each corresponding to an online discussion thread where nodes represent users, and an edge represents the interaction where one of the two users responded to the other user's comment. REDDIT-12K encompasses 11 different categories, with each category corresponding to a different discussion subreddit on Reddit. To create a GOSR benchmark, we partitioned REDDIT-12K and designated the graphs from three of its categories as unseen classes, allowing us to assess the ability of GOSR methods to recognize new, unseen discussion topics.


\subsection{Methods}
\label{sec:Methods}

\begin{table}[t]
    \centering
    \setlength{\abovecaptionskip}{0.cm}
    \caption{Categorization of Node-level and Graph-level GOSR Methods}
    \renewcommand{\arraystretch}{1.5} 
    \scalebox{0.78}{
    \begin{tabular}{c|c|l}
        \toprule
        \textbf{Task Level} & \textbf{Category} & \textbf{Methods} \\ 
        \midrule
        \multirow[c]{9}{*}{\centering Node-level} 
        & \multirow{3}{*}{\centering Traditional OOD \& OSR Methods} 
            & MSP~\cite{DBLP:conf/iclr/HendrycksG17}; Energy~\cite{DBLP:conf/nips/LiuWOL20}\\ 
        & &  ODIN~\cite{DBLP:conf/iclr/HendrycksMD19}; Mah~\cite{DBLP:conf/nips/LeeLLS18}\\ 
        & &  OE~\cite{DBLP:conf/nips/HendrycksMKS19} \\ \cline{2-3}
        & \multirow{2}{*}{\centering GOODD Methods} 
            & GKDE~\cite{zhao2020gkde}; GPN~\cite{stadler2021gpn} \\ 
        & & OODGAT~\cite{huang2022oodgat}; GNNSafe~\cite{wu2023gnnsafe} \\ \cline{2-3}
        & \multirow{1}{*}{\centering GOSR Methods} 
            & EMP~\cite{yang2023emp} \\ \cline{2-3}
        & \multirow{3}{*}{\centering GAD Methods} 
            & GADAR~\cite{ding2022gadar}; GUIDE~\cite{zhu2023guide} \\ 
        & & OCGNN~\cite{wang2021one}; CoLA~\cite{liu2021anomaly} \\ 
        & & CONAD~\cite{luo2022conad}; AnomalyDAE~\cite{ding2019deep} \\ 
        \midrule
        \multirow[c]{6}{*}{\centering Graph-level} 
        & \multirow{2}{*}{\centering Traditional OOD \& OSR Methods} 
            & MSP~\cite{DBLP:conf/iclr/HendrycksG17}; Energy~\cite{DBLP:conf/nips/LiuWOL20} \\ 
        & & Mah~\cite{DBLP:conf/nips/LeeLLS18}; ODIN~\cite{DBLP:conf/iclr/HendrycksMD19} \\ \cline{2-3}
        & \multirow{2}{*}{\centering GOODD Methods} 
            & GraphDE~\cite{li2022graphde}; SGOOD~\cite{ding2023sgood} \\ 
        & & AAGOD~\cite{guo2023aagod} \\ \cline{2-3}
        & \multirow{2}{*}{\centering GAD Methods} 
            & GOOD-D~\cite{liu2023goodd}; OCGIN~\cite{zhang2019ocgin} \\ 
        & & OCGTL~\cite{zhang2020ocgtl}; GlocalKD~\cite{zhang2023glocalkd} \\ 
        \bottomrule
    \end{tabular}}
    \label{tab:gosr_methods_summary_with_refs}
\end{table}

\subsubsection{Node-level Methods}

\paragraph{Traditional OOD \& OSR Methods} 
We provide five traditional OOD and OSR methods. \textbf{MSP}~\cite{DBLP:conf/iclr/HendrycksG17} utilizes the maximum softmax output probability to identify OOD samples. \textbf{Energy}~\cite{DBLP:conf/nips/LiuWOL20} computes energy scores based on the model's output probability distribution for OOD detection. \textbf{ODIN}~\cite{DBLP:conf/iclr/HendrycksMD19} enhances OOD sample detection by introducing small perturbations to the input samples and adjusting the softmax temperature parameter. \textbf{Mah}~\cite{DBLP:conf/nips/LeeLLS18} employs Mahalanobis distance for OOD detection. \textbf{OE}~\cite{DBLP:conf/nips/HendrycksMKS19} trains the model with additional OOD data to improve detection capabilities.

\paragraph{GOODD Methods}  
We provide four novel methods specifically designed for GOODD. \textbf{GKDE}~\cite{zhao2020gkde} identifies OOD nodes by predicting the Dirichlet distribution of nodes. \textbf{GPN}~\cite{stadler2021gpn} employs a graph posterior network approach to enhance node-level OOD detection. \textbf{OODGAT}~\cite{huang2022oodgat} utilizes graph attention mechanisms to improve OOD detection in graphs. \textbf{GNNSafe}~\cite{wu2023gnnsafe} enhances OOD estimation through energy-based trust propagation.

\paragraph{GOSR Methods}    
We provide one methods for GOSR. \textbf{EMP}~\cite{yang2023emp} combines entropy propagation with graph structure information to quantify the likelihood of nodes belonging to unknown classes, and employs entropy clustering to automatically distinguish between known and unknown class nodes.

\paragraph{GAD Methods}   
We provide six methods for GAD. \textbf{GADAR}~\cite{ding2022gadar} utilizes attention and reconstruction techniques to detect anomalies. \textbf{GUIDE}~\cite{zhu2023guide} learns differences between nodes and their higher-order structures through graph attention layers, employing reconstruction error for node-level anomaly detection. \textbf{OCGNN}~\cite{wang2021one} employs one-class classification to detect anomalous nodes in graphs. \textbf{CoLA}~\cite{liu2021anomaly} uses contrastive self-supervised learning to identify anomalies in attributed networks. \textbf{CONAD}~\cite{luo2022conad} employs siamese graph neural networks and contrastive loss for anomaly detection. \textbf{AnomalyDAE}~\cite{ding2019deep} integrates structural and attribute autoencoders to learn the interactions between network structure and node attributes, identifying anomalies through reconstruction error.

\subsubsection{Graph-level Methods}

\paragraph{Traditional OOD \& OSR Methods}
We provide four traditional OOD and OSR methods adapted to the graph level: \textbf{MSP}, \textbf{Energy}, \textbf{Mah} and \textbf{ODIN}.

\paragraph{GOODD Methods}
We provide three novel methods specifically designed for OOD detection in graph data. \textbf{GraphDE} performs OOD detection by modeling the generative process of graph data, introducing a variational recognition model to infer environment variables and employing two-component mixed generative models~\cite{li2022graphde}. \textbf{SGOOD} leverages subgraph structures explicitly to learn graph representations, aiding in graph-level OOD classification~\cite{ding2023sgood}. \textbf{AAGOD} enhances the input of graph neural networks with a parametric amplification matrix to distinguish OOD data at the graph level~\cite{guo2023aagod}.

\paragraph{GAD Methods}
We provide four methods for anomaly detection (AD) at the graph level. \textbf{GOOD-D} utilizes a graph contrastive learning framework to detect OOD samples in graphs through hierarchical contrastive learning and disturbance-free graph data augmentation, without the need for ground truth labels~\cite{liu2023goodd}. \textbf{OCGIN} represents early work in graph-level anomaly detection using one-class classification and graph neural network techniques~\cite{zhang2019ocgin}. \textbf{OCGTL} improves anomaly detection performance by drawing on ideas from self-supervised learning and transformation learning~\cite{zhang2020ocgtl}. \textbf{GlocalKD} conducts graph-level anomaly detection by introducing a global and local knowledge distillation framework to capture both local and global graph characteristics~\cite{zhang2023glocalkd}.

\subsection{Evaluation Metrics}
\label{sec:evaluation}

In this study, we employ three primary metrics to evaluate and compare the performance of different models in graph OOD detection tasks: \textbf{AUROC}, \textbf{AUPR}, \textbf{FPR95}, and \textbf{ACC}. \textbf{AUROC} measures the area under the Receiver Operating Characteristic (ROC) curve and is one of the most commonly used performance indicators. It assesses the model's ability to distinguish between seen classes and unseen classes. The AUROC value ranges from 0 to 1, with higher values indicating stronger discriminative power of the model. \textbf{AUPR} measures the area under the precision-recall curve and is an effective metric for evaluating model performance when the class distribution between seen and unseen classes is imbalanced. A high AUPR means that the model maintains high precision while also achieving a high recall rate in distinguishing seen from unseen classes. \textbf{FPR95} is the false positive rate when the true positive rate (recall rate) is 95$\%$. This metric measures how many seen class samples are incorrectly labeled as unseen when the model correctly identifies 95$\%$ of the unseen class samples. \textbf{ACC} measures the model's ability to classify seen classes accurately.

\section{Experiments}\label{experment}We run all the methods that supported by G-OSR, and compare them on the generalized benchmarks, as shown in \hyperlink{tabTraining_sizes1}{Table 3} and \hyperlink{tabTraining_sizes2}{Table 4}. This section mainly explains our systematic implementation and discussion on the results.

\subsection{Implementation Details}
To ensure consistency and comparability across our graph open-set recognition benchmarks, we standardize the experimental setup by using a two-layer Graph Convolutional Network (GCN) as the backbone architecture for all tested models. We uniformly employ the Adam optimizer for training, with a learning rate set at 0.01 and a weight decay of 0.02, selected to optimize training efficiency and model performance. All experiments are conducted under uniform hardware and software conditions, specifically on a system equipped with a single Nvidia RTX 3090 graphics card, to ensure ample computational resources. Additionally, we mitigate the randomness inherent in machine learning experiments by conducting five independent runs for each method and reporting the average outcomes. This rigorous approach guarantees that our benchmark provides a fair and consistent environment for evaluating various graph open-set recognition methods, enabling reliable comparisons and fostering further research and development in graph processing algorithms.

\begin{table*}[htbp]
	\centering
	\caption{Main Results on Node-Level GOSR Benchmark. The table presents detailed results for each method, including AUROC, AUPR, FPR95, and ACC (\%). The ACC metric for anomaly detection methods is displayed as "-" because these methods are unsupervised and do not classify the seen classes. Additionally, "$\dagger$" indicates that the GPN and EMP experiments on the Arxiv dataset exceeded the GPU memory capacity.}
    \hypertarget{tabTraining_sizes1}{}
	\label{tab:Training_sizes1}
    \scalebox{0.89}{
    \begin{tabular}{@{}l|cccc|cccc|cccc} 
		\toprule
		\multirow[c]{3}{*}{Method} &
		\multicolumn{4}{c|}{Cora} &
		\multicolumn{4}{c|}{Coauthor-CS} &
		\multicolumn{4}{c}{Amazon-Photo} \\
		\cmidrule{2-13}
		& \textbf{AUROC}\,$\uparrow$ & \textbf{AUPR}\,$\uparrow$ & \textbf{FPR95}\,$\downarrow$ & \textbf{ACC}\,$\uparrow$ &
		\textbf{AUROC}\,$\uparrow$ & \textbf{AUPR}\,$\uparrow$ & \textbf{FPR95}\,$\downarrow$ & \textbf{ACC}\,$\uparrow$ &
		\textbf{AUROC}\,$\uparrow$ & \textbf{AUPR}\,$\uparrow$ & \textbf{FPR95}\,$\downarrow$ & \textbf{ACC}\,$\uparrow$ \\
		\midrule
		MSP \cite{DBLP:conf/iclr/HendrycksG17} & 91.13 & 78.27 & 38.19  & 87.64  & 93.89 & 97.61 & 29.71  & 85.19  & 93.65 & 90.83 & 26.97 & 88.05  \\
		Mah \cite{DBLP:conf/nips/LeeLLS18}  & 68.45 & 42.29 & 92.15  & 87.52  & 85.16 & 92.84 & 46.65  & 70.29  & 74.29 & 67.57 & 73.84 & 75.12  \\
		ODIN \cite{DBLP:conf/iclr/HendrycksMD19} & 46.71 & 21.08 & 100.00 & 87.48  & 52.31 & 74.79 & 100.00 & 55.14  & 63.41 & 52.15 & 94.73 & 65.15  \\
		OE \cite{DBLP:conf/nips/HendrycksMKS19} & 89.45 & 77.08 & 45.51  & 87.25  & 95.92 & 97.71 & 19.25  & 84.21  & 95.59 & 92.75 & 17.65 & 87.25  \\
		Energy \cite{DBLP:conf/nips/LiuWOL20} & 91.29 & 78.17 & 42.31  & 87.34  & 94.68 & 97.57 & 20.11  & 85.54  & 93.79 & 90.63 & 29.06 & 88.39  \\
		\midrule
		GKDE \cite{zhao2020gkde} & 57.23 & 27.50 & 88.95  & 89.87  & 61.15 & 81.39 & 94.60  & 89.05  & 65.58 & 65.20 & 96.87 & 89.37  \\
		GPN \cite{stadler2021gpn} & 90.21 & 77.32 & 36.34  & 87.17  & 93.51 & 97.53 & 35.31  & 86.08  & 93.42 & 90.61 & 37.06 & 89.13  \\
		OODGAT \cite{huang2022oodgat} & 91.81 & 80.74 & 36.34  & 87.05  & 95.36 & 98.16 & 24.58  & 87.09  & 94.23 & 91.85 & 23.05 & 90.03  \\
        GNNSafe \cite{wu2023gnnsafe} & 92.07 & 81.45 & 34.66  & 87.43  & 97.81 & 99.19 & 9.78   & 88.07  & 97.32 & 96.94 & 6.21  & 90.99  \\
		\midrule
        EMP \cite{yang2023emp} & 95.74 & 85.34 & 18.12  & 90.26  & 98.44 & 99.28 & 6.18   & 92.93  & 98.34 & 95.19 & 10.16 & 89.84  \\
		\midrule
        AnomalyDAE \cite{ding2019deep} & 48.19 & 65.34 & 96.66   & -      & 62.70 & 37.75 & 93.55  & -      & 51.72 & 60.08 & 95.28 & -     \\
		OCGNN \cite{wang2021one} & 48.58 & 67.11 & 95.66  & -      & 51.25 & 28.01 & 96.05  & -      & 46.63 & 55.91 & 95.67 & -    \\
		CoLA \cite{liu2021anomaly} & 58.47 & 72.66 & 94.32  & -      & 42.21 & 22.95 & 96.48  & -      & 51.41 & 61.06 & 95.16 & -    \\
        GADNR \cite{ding2022gadar} & 57.63 & 73.83 & 94.88  & -      & 50.93 & 28.05 & 96.23  & -      & 42.19 & 54.18 & 99.14 & -        \\
		CONAD \cite{luo2022conad} & 49.77 & 67.04 & 95.22  & -      & 63.71 & 80.51 & 95.57  & -      & 64.91 & 65.28 & 97.13 & -    \\
		GUIDE \cite{zhu2023guide} & 55.88 & 71.11 & 94.01  & -      & 59.98 & 74.98 & 89.98  & -      & 54.98 & 69.98 & 89.98 & -    \\
		\midrule \midrule
        \multirow[c]{3}{*}{Method} & \multicolumn{4}{c|}{Coauthor-Physics} & \multicolumn{4}{c|}{Citeseer} & \multicolumn{4}{c}{Arxiv} \\
		\cmidrule{2-13}
		& 
		\textbf{AUROC}\,$\uparrow$ & \textbf{AUPR}\,$\uparrow$ & \textbf{FPR95}\,$\downarrow$ & \textbf{ACC}\,$\uparrow$ &
		\textbf{AUROC}\,$\uparrow$ & \textbf{AUPR}\,$\uparrow$ & \textbf{FPR95}\,$\downarrow$ & \textbf{ACC}\,$\uparrow$ &
		\textbf{AUROC}\,$\uparrow$ & \textbf{AUPR}\,$\uparrow$ & \textbf{FPR95}\,$\downarrow$ & \textbf{ACC}\,$\uparrow$ \\
		\midrule
		MSP \cite{DBLP:conf/iclr/HendrycksG17} & 93.95 & 98.82 & 28.67 & 97.78  & 86.11 & 95.12 & 57.44 & 68.49 & 68.63 & 80.70 & 89.24 & 70.14 \\
		Mah \cite{DBLP:conf/nips/LeeLLS18} & 54.15 & 88.72 & 96.34 & 65.16  & 58.25 & 82.35 & 92.93 & 68.31 & 56.94 & 69.36 & 94.27 & 65.13 \\
		ODIN \cite{DBLP:conf/iclr/HendrycksMD19} & 43.77 & 88.52 & 99.72 & 65.65  & 13.94 & 60.55 & 99.59 & 68.53 & 39.69 & 59.01 & 99.19 & 50.14 \\
		OE \cite{DBLP:conf/nips/HendrycksMKS19} & 97.14 & 99.43 & 12.64 & 98.04  & 90.10 & 96.61 & 47.62 & 68.73 & 69.53 & 79.46 & 86.73 & 69.36 \\
		Energy \cite{DBLP:conf/nips/LiuWOL20} & 99.32 & 99.87 & 2.76  & 98.06  & 90.49 & 97.49 & 87.64 & 68.56 & 64.75 & 76.25 & 90.16 & 69.13 \\
		\midrule
		GKDE \cite{zhao2020gkde} & 60.41 & 80.57 & 95.20 & 89.21  & 60.73 & 85.12 & 90.48 & 88.99 & 54.94 & 72.32 & 88.65 & 68.57 \\
		GPN \cite{stadler2021gpn} & 94.98 & 96.97 & 32.02 & 97.07  & 87.98 & 94.97 & 52.02 & 68.13 & $\dagger$ & $\dagger$ & $\dagger$ & $\dagger$ \\
		OODGAT \cite{huang2022oodgat} & 94.97 & 96.97 & 21.97 & 97.04  & 88.03 & 95.96 & 45.02 & 68.15 & 62.71 & 76.19 & 85.37 & 70.41 \\
        GNNSafe \cite{wu2023gnnsafe} & 99.32 & 99.86 & 2.75  & 98.06  & 90.49 & 97.49 & 87.57 & 68.79 & 59.81 & 54.23 & 84.71 & 69.55 \\
		\midrule
        EMP \cite{yang2023emp} & 98.62 & 99.65 & 5.47  & 98.15  & 88.18 & 95.24 & 54.66 & 74.84 & $\dagger$ & $\dagger$ & $\dagger$ & $\dagger$ \\
		\midrule
        AnomalyDAE \cite{ding2019deep} & 50.08 & 65.17 & 95.02 & -      & 55.04 & 69.89 & 90.11 & -      & 49.43 & 64.27 & 94.79 & - \\
		OCGNN \cite{wang2021one} & 50.45 & 65.03 & 94.88 & -      & 54.89 & 70.31 & 89.82 & -      & 48.75 & 63.26 & 93.96 & - \\
		CoLA \cite{liu2021anomaly} & 45.22 & 60.14 & 94.77 & -      & 49.85 & 64.91 & 89.98 & -      & 46.47 & 60.81 & 95.14 & - \\
        GADNR \cite{ding2022gadar} & 49.96 & 59.87 & 94.73 & -      & 55.38 & 70.22 & 90.64 & -      & 47.71 & 60.88 & 94.52 & - \\
		CONAD \cite{luo2022conad} & 54.82 & 69.76 & 89.93 & -      & 60.34 & 75.07 & 89.85 & -      & 48.23 & 62.34 & 92.99 & - \\
		GUIDE \cite{zhu2023guide} & 60.25 & 75.32 & 90.24 & -      & 59.77 & 74.53 & 89.71 & -      & 46.62 & 61.85 & 92.87 & - \\
        \bottomrule
	\end{tabular}}

\end{table*}

\subsection{ Main Results}

\subsubsection{Analysis of Main Results}

\paragraph{\textbf{Auxiliary data significantly enhances the detection of unseen classes.}}
In node-level tasks (\hyperlink{tabTraining_sizes1}{Table 3}), experimental results indicate that integrating additional data can significantly enhance the performance of unseen class detection methods on node-level GOSR tasks. For instance, on the Coauthor-CS dataset, the OE method achieves an AUROC of 95.92\%, which is 2.03\% higher than MSP (AUROC of 93.89\%). Additionally, OE's FPR95 is 19.25\%, which is significantly lower than MSP's 29.71\%, indicating that OE incorrectly labels fewer seen class samples as unseen. This improvement highlights the value of external data sources in improving the model's ability to identify unseen class samples, particularly when there is a significant disparity between the distributions of seen and unseen classes.

\paragraph{\textbf{Post-hoc methods exhibit significant limitations in tasks at both levels}}
In node-level tasks (\hyperlink{tabTraining_sizes1}{Table 3}), based on the experimental results, post-hoc methods such as ODIN and Mahalanobis distance exhibit significant limitations when applied to node-level GOSR tasks. For example, on the Cora dataset, Mah achieves an AUROC of 68.45\%, which is significantly lower than GNNSafe's 92.07\%. Moreover, Mah's FPR95 is 92.15\%, compared to GNNSafe's 34.66\%, showing that Mah struggles to accurately detect unseen classes. These methods show poor performance with low AUROC and AUPR values and high FPR95 scores, particularly when dealing with complex structures and diverse node characteristics inherent in graph data.

In graph-level GOSR tasks (\hyperlink{tabTraining_sizes2}{Table 4}), experimental results indicate that Post-hoc methods, such as MSP, Mahalanobis distance, ODIN, and Energy, generally perform poorly, especially when dealing with complex graph datasets. For instance, on the ENZYMES dataset, the AUROC scores of these Post-hoc methods range from 60.86\% to 67.56\%. Although Mahalanobis distance performs slightly better than other Post-hoc methods, it still falls short compared to the graph-specific method SGOOD, which achieves an AUROC of 74.52\%. The performance of Post-hoc methods deteriorates further on the more challenging IMDB-M dataset, with Energy achieving a particularly low AUROC of just 25.45\%, indicating its difficulty in effectively capturing the features of complex graph data. While Post-hoc methods show some improvement in AUROC on the BBBP dataset, they still lag significantly behind graph-specific methods like AAGOD. Graph-specific methods, such as SGOOD and AAGOD, consistently demonstrate superior classification ability and lower false positive rates across all datasets, highlighting their advantage in capturing the structural information inherent in graphs, and thus, outperforming in graph-level GOSR tasks.

\paragraph{\textbf{Graph Anomaly Detection methods seem not well-suited for GOSR tasks}}
In node-level tasks (\hyperlink{tabTraining_sizes1}{Table 3}), Graph Anomaly Detection (GAD) methods typically focus on detecting local anomalies in the structure or attributes of graphs. While this approach may be effective for identifying certain types of outliers, it performs poorly in node-level GOSR tasks. For instance, on the Coauthor-CS dataset, GUIDE achieves an AUROC of only 59.98\%, significantly lower than EMP's 98.44\%. GUIDE's FPR95 is 89.98\%, compared to EMP's 6.18\%. This suggests that GAD methods underperform in distinguishing unseen classes from seen classes. The reason behind this phenomenon might be that GAD methods tend to focus on local structural differences within the graph rather than on global features. When dealing with large-scale and complex graph data, GAD methods may struggle to capture the overall differences between nodes of unseen and seen classes. Moreover, GAD methods are typically not designed to handle OSR tasks, limiting their detection capability in open environments.

In the experimental results of graph-level tasks (\hyperlink{tabTraining_sizes2}{Table 4}), the performance of graph anomaly detection methods varied significantly across different datasets. On the ENZYMES dataset, these methods performed relatively well, with AUROC values ranging between 65\% and 68\%, indicating their capability in identifying anomalous graph structures. However, on more complex datasets such as IMDB-M and BBBP, the AUROC values for these methods were considerably lower, making it difficult to effectively distinguish unseen classes. For instance, GlocalKD achieved an AUROC of only 21.67\% on the IMDB-M dataset, while GOOD-D recorded an AUROC of 52.95\% on the BBBP dataset, both of which are significantly lower than the performance of graph OOD methods. Overall, while graph anomaly detection methods can capture local structural anomalies in certain datasets, they fall short when handling more complex graph data. In contrast, graph OOD methods like SGOOD and AAGOD consistently demonstrate superior performance, particularly in terms of classification accuracy and false positive rates, underscoring the limitations of graph anomaly detection methods in graph-level GOSR tasks, especially when dealing with intricate graph structures.

\paragraph{\textbf{Node and graph complexity significantly degrades method performance on complex datasets}}
Node feature complexity and graph topology complexity significantly affect the performance of different methods in node-level GOSR tasks. This impact becomes evident when comparing datasets like Cora and Arxiv. As shown in \hyperlink{tabTraining_sizes1}{Table 3}, in the Cora dataset, GNNSafe and EMP achieve AUROCs of 92.07\% and 95.74\%, respectively, while traditional methods like MSP, OE, and Energy also reach AUROCs around 90\%. This indicates that when node features and graph topology are relatively simple, the performance differences among these methods are minimal, and all can adequately handle the GOSR tasks. However, in the Arxiv dataset, where node feature complexity and graph topology are more intricate, the performance drops significantly. MSP achieves an AUROC of 68.63\%, and GNNSafe's AUROC drops to 59.81\%, both substantially lower than their performance on the simpler Cora dataset. This significant decline demonstrates that as the complexity of node features and graph structure increases, the ability of both traditional and graph-specific methods to address GOSR tasks diminishes. This underscores the evident limitations of existing methods when applied to node-level GOSR tasks in more complex graph data.

The experimental results indicate that node feature complexity and graph topology complexity also impact the performance of methods in graph-level GOSR tasks (\hyperlink{tabTraining_sizes2}{Table 4}). For datasets with relatively simple structures, such as ENZYMES, both graph-specific methods (e.g., SGOOD, AAGOD) and anomaly detection methods achieve relatively high AUROC scores, suggesting that the effect of complexity is minimal. However, as complexity increases, and particularly in highly complex datasets such as REDDIT-12K and IMDB-M, the performance of traditional methods and anomaly detection methods declines significantly. For instance, Energy achieves an AUROC of only 25.45\% on IMDB-M, highlighting its struggle with complex graph data. In contrast, graph-specific methods maintain strong performance even on challenging datasets, showcasing their superior robustness and classification abilities.

\begin{table*}[htbp]
	\centering
	\caption{Main Results on Graph-Level GOSR Benchmark. The table presents detailed results for each method, including AUROC, AUPR, FPR95, and ACC (\%). The ACC metric for anomaly detection methods is displayed as "-" because these methods are unsupervised and do not classify the seen classes.}
    \hypertarget{tabTraining_sizes2}{}
	\label{tab:Training_sizes2}
    \scalebox{0.93}{
	\begin{tabular}{@{}l|cccc|cccc|cccc@{}} 
		\toprule
		\multirow[c]{3}{*}{Method} &
		\multicolumn{4}{c|}{ENZYMES} &
		\multicolumn{4}{c|}{IMDB-M} &
		\multicolumn{4}{c}{BBBP} \\
		\cmidrule{2-13}
		& \textbf{AUROC}\,$\uparrow$ & \textbf{AUPR}\,$\uparrow$ & \textbf{FPR95}\,$\downarrow$ & \textbf{ACC}\,$\uparrow$ &
		  \textbf{AUROC}\,$\uparrow$ & \textbf{AUPR}\,$\uparrow$ & \textbf{FPR95}\,$\downarrow$ & \textbf{ACC}\,$\uparrow$ &
		  \textbf{AUROC}\,$\uparrow$ & \textbf{AUPR}\,$\uparrow$ & \textbf{FPR95}\,$\downarrow$ & \textbf{ACC}\,$\uparrow$ \\
		\midrule
		MSP \cite{DBLP:conf/iclr/HendrycksG17} & 60.86 & 61.48 & 90.76 & 42.37 & 43.65 & 51.76 & 94.38 & 50.45 & 54.96 & 57.03 & 95.72 & 88.98 \\
		Mah \cite{DBLP:conf/nips/LeeLLS18} & 67.56 & 63.41 & 84.05 & 41.93 & 70.39 & 64.63 & 58.98 & 50.02 & 53.85 & 52.37 & 93.22 & 88.23 \\
		ODIN \cite{DBLP:conf/iclr/HendrycksMD19} & 63.26 & 66.03 & 92.94 & 42.08 & 40.46 & 50.75 & 96.31 & 50.65 & 55.36 & 55.27 & 95.93 & 88.47 \\
		Energy \cite{DBLP:conf/nips/LiuWOL20} & 56.74 & 58.38 & 88.56 & 43.76 & 25.45 & 38.43 & 96.16 & 50.73 & 55.94 & 55.13 & 92.85 & 89.03 \\
        \midrule
        GraphDE \cite{li2022graphde} & 61.75 & 65.83 & 99.01 & 48.67 & 67.28 & 63.12 & 92.47 & 42.38 & 50.69 & 51.38 & 94.75 & 89.58 \\
		SGOOD \cite{ding2023sgood} & 74.52 & 71.45 & 71.00 & 51.16 & 75.99 & 68.45 & 51.60 & 51.08 & 60.09 & 58.40 & 92.44 & 90.16 \\
		AAGOD \cite{guo2023aagod} & 72.86 & 75.08 & 87.34 & 43.97 & 65.86 & 67.08 & 87.34 & 49.83 & 67.86 & 61.08 & 90.34 & 88.54 \\
        \midrule
        OCGIN \cite{zhang2019ocgin} & 63.25 & 68.11 & 91.14 & - & 50.05 & 55.42 & 93.82 & - & 45.12 & 48.45 & 95.01 & - \\
        OGGTL \cite{zhang2020ocgtl} & 68.12 & 70.45 & 89.93 & - & 54.34 & 57.59 & 85.22 & - & 48.25 & 51.03 & 9.34 & - \\
        GLocalKD \cite{zhang2023glocalkd} & 65.48 & 69.87 & 90.23 & - & 21.67 & 36.31 & 92.14 & - & 57.36 & 49.94 & 93.12 & - \\
        GOOD-D \cite{liu2023goodd} & 65.79 & 66.23 & 92.34 & - & 68.96 & 70.12 & 98.75 & - & 52.95 & 50.53 & 92.45 & - \\
		\midrule \midrule
		\multirow[c]{3}{*}{Method} &
        \multicolumn{4}{c|}{BZR} &
        \multicolumn{4}{c|}{Tox21} &
        \multicolumn{4}{c}{REDDIT-12K} \\
		\cmidrule{2-13}
		& \textbf{AUROC}\,$\uparrow$ & \textbf{AUPR}\,$\uparrow$ & \textbf{FPR95}\,$\downarrow$ & \textbf{ACC}\,$\uparrow$ &
		  \textbf{AUROC}\,$\uparrow$ & \textbf{AUPR}\,$\uparrow$ & \textbf{FPR95}\,$\downarrow$ & \textbf{ACC}\,$\uparrow$ &
		  \textbf{AUROC}\,$\uparrow$ & \textbf{AUPR}\,$\uparrow$ & \textbf{FPR95}\,$\downarrow$ & \textbf{ACC}\,$\uparrow$ \\
		\midrule
		MSP \cite{DBLP:conf/iclr/HendrycksG17} & 72.32 & 74.24 & 38.64 & 79.58 & 62.12 & 67.89 & 93.12 & 75.14 & 58.74 & 60.32 & 90.74 & 49.23 \\
		Mah \cite{DBLP:conf/nips/LeeLLS18} & 73.11 & 75.36 & 45.78 & 76.89 & 60.45 & 69.12 & 89.65 & 78.54 & 70.72 & 73.84 & 82.62 & 49.82 \\
		ODIN \cite{DBLP:conf/iclr/HendrycksMD19} & 71.31 & 73.74 & 51.35 & 76.12 & 63.89 & 66.54 & 95.12 & 72.09 & 59.67 & 61.32 & 91.67 & 48.97 \\
		Energy \cite{DBLP:conf/nips/LiuWOL20} & 70.46 & 72.35& 39.24 & 78.01 & 62.93 & 65.87 & 91.35 & 73.45 & 57.23 & 59.12 & 91.23 & 49.39 \\
        \midrule
        GraphDE \cite{li2022graphde} & 69.89 & 66.12 & 57.98 & 77.54 & 61.23 & 68.47 & 90.12 & 74.91 & 58.12 & 60.54 & 89.32 & 44.49 \\
		SGOOD \cite{ding2023sgood} & 74.34 & 72.45 & 57.89 & 81.21 & 66.45 & 72.98 & 80.89 & 81.12 & 72.81 & 76.17 & 79.72 & 51.23 \\
		AAGOD \cite{guo2023aagod} & 81.54 & 83.67 & 35.43 & 79.12 & 65.32 & 71.89 & 88.45 & 76.53 & 70.67 & 68.28 & 85.57 & 47.45 \\
        \midrule
        OCGIN \cite{zhang2019ocgin} & 72.12 & 77.21 & 59.34 & - & 52.35 & 60.21 & 80.98 & - & 58.23 & 57.36 & 93.67 & - \\
        OGGTL \cite{zhang2020ocgtl} & 65.12 & 69.87 & 52.54 & - & 49.52 & 52.68 & 79.02 & - & 53.54 & 55.67 & 92.34 & - \\
        GLocalKD \cite{zhang2023glocalkd} & 53.78 & 58.12 & 69.01 & - & 42.53 & 50.15 & 82.12 & - & 47.18 & 56.15 & 96.01 & - \\
        GOOD-D \cite{liu2023goodd} & 70.98 & 74.89 & 50.12 & - & 60.56 & 68.45 & 81.45 & - & 58.29 & 61.28 & 90.78 & - \\
		\bottomrule
	\end{tabular}}
\end{table*}

\paragraph{\textbf{Graph-specific methods generally outperform traditional methods in GOSR, though challenges remain with more complex data}}
In node-level GOSR tasks, traditional methods such as MSP, Mahalanobis distance, ODIN, and energy-based approaches generally underperform compared to graph-specific methods. For example, on the Cora dataset (\hyperlink{tabTraining_sizes1}{Table 3}), MSP achieves an AUROC of 91.13\%, while Mahalanobis distance, ODIN, and energy-based methods score lower, with AUROCs of 68.45\%, 46.71\%, and 91.29\%, respectively. The FPR95 values further highlight the limitations of these traditional methods, with ODIN reaching 100\%, indicating a high false positive rate when distinguishing unseen classes from seen ones. In contrast, graph-specific methods generally perform better in node-level tasks. For instance, GNNSafe achieves an AUROC of 92.07\% on the Cora dataset, with an FPR95 of 34.66\%, significantly outperforming traditional methods. Similarly, EMP achieves the highest AUROC of 95.74\% and the lowest FPR95 of 18.12\% across the three datasets analyzed (Cora, Coauthor-CS, and Amazon-Photo).
However, on the Arxiv dataset (\hyperlink{tabTraining_sizes1}{Table 3}), graph-specific methods like GNNSafe (AUROC 59.81\%) do not show a significant advantage over traditional methods such as MSP (AUROC 68.63\%), OE (AUROC 69.53\%), and Energy (AUROC 64.75\%). In fact, they sometimes underperform. This suggests that current node-level graph-specific GOSR methods may be better suited for simpler or moderately complex graph structures, but their ability to extract effective features may be insufficient when dealing with more complex, large-scale datasets. Therefore, future research should focus on further optimizing graph-specific methods to ensure they perform well on more complex graph data.

In graph-level GOSR tasks, traditional methods continue to underperform compared to graph-specific methods. For example, on the BBBP dataset (\hyperlink{tabTraining_sizes2}{Table 4}), the AUROC of MSP is 54.96\%, while Mahalanobis distance and ODIN score similarly low at 53.85\% and 55.36\%, respectively. These methods also exhibit high FPR95 values, exceeding 90\% in many cases, indicating their limited ability to accurately classify unseen classes. In contrast, graph-specific methods such as SGOOD and AAGOD show much stronger performance. SGOOD achieves an AUROC of 60.09\% on the BBBP dataset, while AAGOD scores 67.86\% on the same dataset. On the IMDB-M dataset, SGOOD achieves an AUROC of 75.99\%, outperforming traditional methods by a significant margin. Furthermore, the FPR95 values for these methods are substantially lower than those of traditional methods, reinforcing their effectiveness in graph-level GOSR tasks.

\begin{table*}[htbp]
    \centering
    \caption{Performance Comparison of Various Methods on the CoraFull Dataset Across Different Numbers of Seen Classes. The table presents detailed results for each method, including AUROC, AUPR, FPR95, and ACC (\%), measured under varying numbers of seen classes (5, 10, 20, 30, and 40).}
    \hypertarget{tabcorafull_results}{}
    \label{tab:corafull_results}
    \scalebox{0.93}{
    \begin{tabular}{@{}l|cccc|cccc|cccc@{}}
        \toprule
        \textbf{Method} & \multicolumn{4}{c|}{\textbf{5 Seen Classes}} & \multicolumn{4}{c|}{\textbf{10 Seen Classes}} & \multicolumn{4}{c}{\textbf{20 Seen Classes}}  \\ 
        \midrule
        & \textbf{AUROC}\,$\uparrow$ & \textbf{AUPR}\,$\uparrow$ & \textbf{FPR95}\,$\downarrow$ & \textbf{ACC}\,$\uparrow$ &
        \textbf{AUROC}\,$\uparrow$ & \textbf{AUPR}\,$\uparrow$ & \textbf{FPR95}\,$\downarrow$ & \textbf{ACC}\,$\uparrow$ &
        \textbf{AUROC}\,$\uparrow$ & \textbf{AUPR}\,$\uparrow$ & \textbf{FPR95}\,$\downarrow$ & \textbf{ACC}\,$\uparrow$ \\
        \midrule
        MSP \cite{DBLP:conf/iclr/HendrycksG17}      & 97.93 & 97.11 & 7.34 & 91.07   & 92.07 & 94.48 & 26.95 & 89.15   & 85.62 & 94.22 & 53.82 & 80.43  \\
        Energy \cite{DBLP:conf/nips/LiuWOL20}   & 96.02 & 96.45 & 8.12 & 90.23   & 91.89 & 93.68 & 29.22 & 88.91   & 84.58 & 93.74 & 55.34 & 79.98  \\
        Mah \cite{DBLP:conf/nips/LeeLLS18}     & 68.45 & 72.18 & 97.34 & 67.12  & 65.78 & 70.45 & 98.12 & 66.23   & 61.23 & 71.67 & 97.12 & 65.78  \\
        ODIN \cite{DBLP:conf/iclr/HendrycksMD19}     & 57.12 & 70.78 & 98.75 & 64.89  & 55.34 & 72.11 & 98.93 & 63.57   & 50.67 & 70.11 & 99.12 & 64.89  \\
        OE \cite{DBLP:conf/nips/HendrycksMKS19}      & 98.16 & 97.65 & 6.41 & 90.31   & 93.32 & 95.27 & 25.64 & 89.57   & 87.63 & 95.08 & 50.81 & 80.33 \\
        \midrule
        GKDE \cite{zhao2020gkde}     & 67.89 & 68.12 & 89.34 & 71.54  & 65.23 & 66.87 & 91.12 & 70.12   & 63.45 & 66.78 & 92.12 & 69.12 \\
        GPN \cite{stadler2021gpn}      & 85.97 & 94.33 & 82.41 & 88.79   & 84.54 & 95.12 & 83.02 & 88.32   & 85.89 & 96.21 & 82.65 & 88.12  \\
        OODGAT \cite{huang2022oodgat}   & 88.12 & 94.28 & 81.89 & 88.94   & 88.67 & 95.01 & 82.76 & 88.54   & 87.93 & 96.04 & 83.01 & 88.32  \\
        GNNSafe \cite{wu2023gnnsafe}  & 89.35 & 92.32 & 80.31 & 90.57  & 89.69 & 95.22 & 82.24 & 89.49   & 87.66 & 96.45 & 83.78 & 80.58 \\
        \midrule
        EMP \cite{yang2023emp}      & 89.31 & 92.75 & 77.14  & 90.63   & 90.12 & 95.58 & 69.85 & 90.28   & 90.34 & 96.12 & 68.14 & 90.12  \\
        \midrule \midrule
        \textbf{Method} & \multicolumn{4}{c|}{\textbf{30 Seen Classes}} & \multicolumn{4}{c|}{\textbf{40 Seen Classes}} & \multicolumn{3}{c}{\textbf{Robustness Metrics}} \\ 
        \midrule
        & \textbf{AUROC}\,$\uparrow$ & \textbf{AUPR}\,$\uparrow$ & \textbf{FPR95}\,$\downarrow$ & \textbf{ACC}\,$\uparrow$ &
        \textbf{AUROC}\,$\uparrow$ & \textbf{AUPR}\,$\uparrow$ & \textbf{FPR95}\,$\downarrow$ & \textbf{ACC}\,$\uparrow$ & \textbf{SD}\,$\downarrow$ & \textbf{MD}\,$\downarrow$ & \textbf{RS}\,$\uparrow$ \\
        \midrule
        MSP \cite{DBLP:conf/iclr/HendrycksG17}     & 82.84 & 95.64 & 59.00 & 74.05   & 82.19 & 96.70 & 60.62 & 69.87  & 6.72 & 15.74 & -4.07\\
        Energy \cite{DBLP:conf/nips/LiuWOL20}   & 81.92 & 94.87 & 60.12 & 73.98   & 81.32 & 95.92 & 61.34 & 68.79   & 6.50 & 14.70 & -3.93 \\
        Mah \cite{DBLP:conf/nips/LeeLLS18}      & 60.74 & 72.18 & 98.37 & 63.62  & 58.67 & 70.12 & 99.21 & 62.11  & 4.01 & 9.78 & -2.46\\
        ODIN \cite{DBLP:conf/iclr/HendrycksMD19}     & 45.09 & 74.11 & 99.15 & 68.16  & 43.98 & 72.88 & 99.45 & 65.87  & 5.89 & 13.14 & -3.65\\
        OE \cite{DBLP:conf/nips/HendrycksMKS19}       & 84.55 & 95.99 & 52.59 & 74.40  & 83.02 & 96.81 & 55.52 & 70.10  & 6.31 & 15.14 & -3.90 \\
        \midrule
        GKDE \cite{zhao2020gkde}    & 61.12 & 65.78 & 94.11 & 68.56  & 59.87 & 64.11 & 96.89 & 68.23  & 3.20 & 8.02 & -2.01\\
        GPN \cite{stadler2021gpn}    & 83.45 & 97.02 & 80.23 & 89.45  & 83.12 & 97.78 & 79.78 & 89.02  & 1.32 & 2.84 & -0.67\\
        OODGAT \cite{huang2022oodgat}  & 89.78 & 96.89 & 78.56 & 89.34  & 89.45 & 97.23 & 77.89 & 88.78   & 0.80 & 1.84 & 0.37 \\
        GNNSafe \cite{wu2023gnnsafe} & 89.14 & 97.95 & 81.47 & 74.58  & 90.78 & 98.62 & 67.34 & 70.02   & 1.12 & 3.12 & 0.23\\
        \midrule
        EMP \cite{yang2023emp}     & 87.98 & 96.12 & 69.65  & 89.54  & 88.02 & 97.23 & 60.12 & 89.12   & 1.12 & 2.35 & -0.47\\
        \bottomrule
    \end{tabular}}
\end{table*}

\begin{figure}[htbp]
    \centering
    \includegraphics[width=\linewidth]{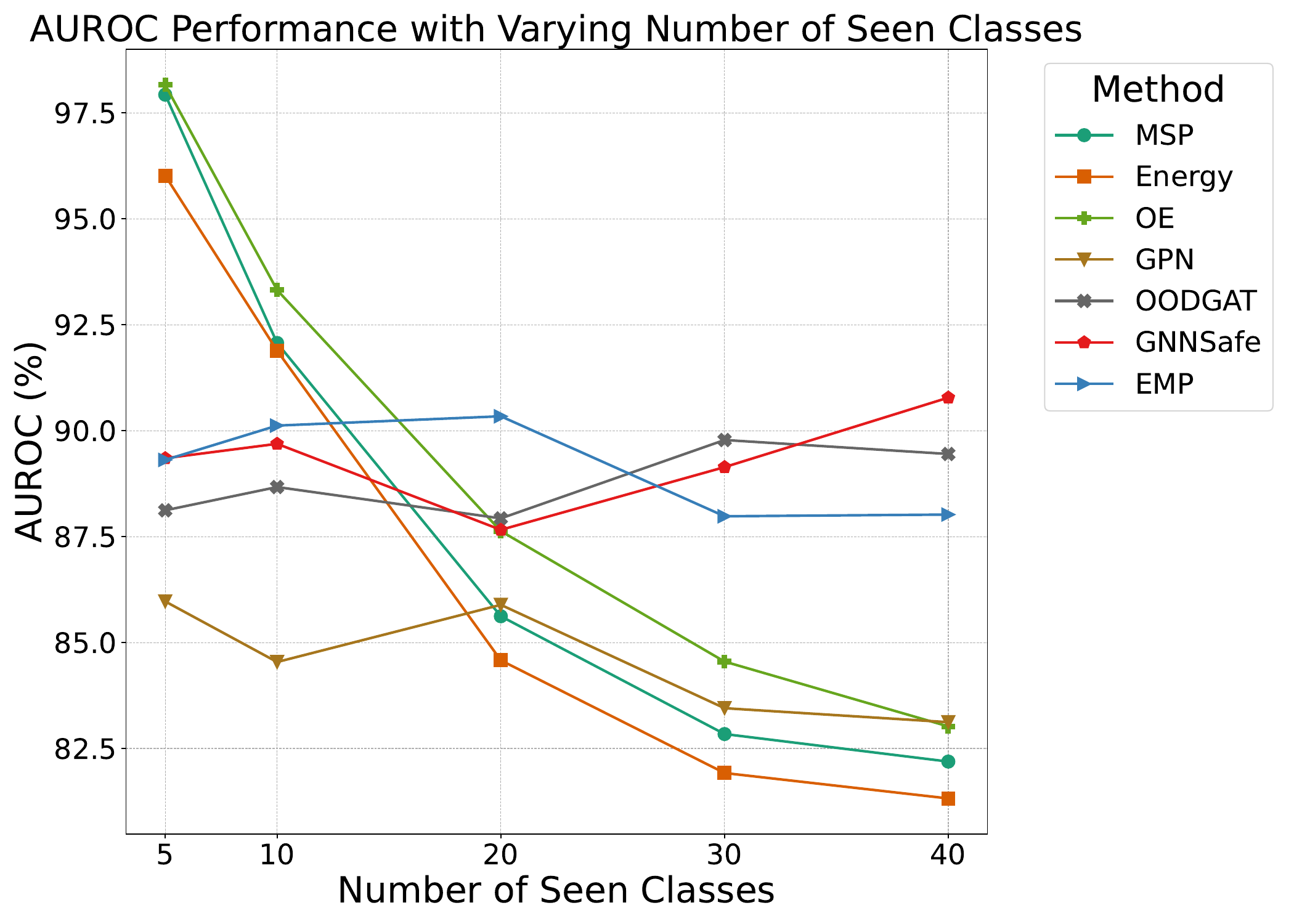}
    \caption{AUROC Performance with Varying Number of Seen Classes.}
    \hypertarget{figpic1}{}
    \hypertarget{figacc_performance}{}
    \label{fig:acc_performance}
\end{figure}

\subsubsection{Robustness Analysis}

\paragraph{\textbf{Experiments Setup}}
In addition to our main experiments, we conducted additional tests to evaluate the robustness of different methods when faced with variations in the class space. Specifically, we used the CoraFull dataset, a node classification dataset containing 70 classes, to examine how well each method performs as the number of seen and unseen classes changes. In these experiments, we systematically selected different numbers of classes as seen classes, while the remaining classes were treated as unseen. By varying the number of seen classes, we were able to analyze the adaptability and stability of each method in scenarios where the complexity of the task increases due to a growing number of categories. This section provides a detailed analysis of the experimental results, highlighting the strengths and weaknesses of each method under these varying conditions. In this part of our evaluation, we utilized three metrics to assess the performance of the methods: \textbf{Standard Deviation (SD)}, \textbf{Regression Slope (RS)}, and \textbf{Maximum Difference (MD)}. \textbf{SD} measures the variability in performance across different class setups, indicating how consistent a method is—a lower value suggests better robustness. \textbf{RS} determines the trend of performance changes with increasing numbers of seen classes, helping us understand each method's adaptability to new class information—a positive or near-zero slope suggests stable and robust performance. \textbf{MD} calculates the maximum difference in performance across different class setups, highlighting the worst-case scenario for robustness—a lower value indicates higher robustness. These metrics collectively provide insights into the methods' consistency, stability, and reliability under varying class configurations. Based on the experimental results of different methods under varying class spaces shown in \hyperlink{tabcorafull_results}{Table 5}, and the performance trends depicted in \hyperlink{figacc_performance}{Figure 2}, we draw some interesting conclusions.

\paragraph{\textbf{Traditional methods show poor robustness as class complexity increases}}
The experimental results reveal that traditional open-set recognition methods (such as MSP, Energy, Mah, and ODIN) perform relatively well when the number of seen classes is small (e.g., 5 Seen Classes), often achieving higher AUROC and ACC values. For instance, MSP achieves an AUROC of 97.93\% and an ACC of 91.07\% with 5 seen classes, while Energy records an AUROC of 96.02\% and an ACC of 90.23\%. However, as the number of seen classes increases, the performance of these methods significantly deteriorates. Specifically, MSP’s AUROC drops to 82.19\% and ACC to 69.87\% with 40 seen classes, and Energy’s AUROC declines to 81.32\% with an ACC of 68.79\%. This performance degradation is reflected in their higher standard deviation (SD) and maximum difference (MD), where MSP has an SD of 6.72 and an MD of 15.74, and Energy shows similar values with an SD of 6.50 and an MD of 14.70.
Mah and ODIN methods exhibit particularly unstable performance across all experimental splits. Mah's AUROC decreases from 68.45\% with 5 seen classes to 58.67\% with 40 seen classes, while ODIN's AUROC falls from 57.12\% to 43.98\%. Notably, ODIN’s FPR95 rises to 99.45\% at 40 seen classes, indicating a significantly reduced capability to handle false positives. The negative regression slope (RS) for Mah (-2.46) and ODIN (-3.65) further illustrates the pronounced decline in performance as the number of seen classes increases.
This phenomenon can be attributed to the inherent limitations of traditional methods in handling high-dimensional and complex data. As the number of classes increases, the decision boundaries between classes become more complex. Traditional methods, which often rely on fixed thresholds or predefined distributions, struggle to accurately distinguish between seen and unseen classes in high-dimensional, multi-class scenarios. This leads to a decrease in their performance as the class space expands.

\paragraph{\textbf{Graph-specific methods maintain better robustness with increasing class complexity}}
In contrast to traditional methods, graph-specific methods (such as GPN, OODGAT, GNNSafe, and EMP) exhibit significantly better robustness when dealing with an increasing number of classes. For example, GPN maintains a relatively stable AUROC, starting at 85.97\% with 5 seen classes and slightly decreasing to 83.12\% with 40 seen classes. Similarly, GNNSafe's AUROC remains consistently high, starting at 89.35\% with 5 seen classes and increasing to 90.78\% with 40 seen classes. EMP also shows minimal fluctuation, with AUROC values ranging from 89.31\% to 88.02\% across different splits.
The low SD and MD values for these graph-specific methods further confirm their robustness across varying class spaces. For instance, GNNSafe has an SD of 1.12 and an MD of 3.12, while EMP exhibits an SD of 1.12 and an MD of 2.35. In contrast, traditional methods exhibit much higher SD and MD, indicating greater performance volatility.

The superior robustness of graph-specific methods can be attributed to their ability to leverage graph structure and the relationships between nodes to capture more complex patterns. These methods, which utilize graph neural networks, are better equipped to adapt to changes in the class space and maintain consistent performance across different scenarios. Additionally, graph-specific methods excel in handling scenarios with blurred class boundaries, likely due to their capacity to extract richer contextual information from the graph structure, thereby enhancing their adaptability to complex environments.

\paragraph{\textbf{Graph-specific methods show no advantage over traditional methods when the number of seen classes is small, but outperform them as the number of seen classes increases}}
The experimental results suggest an interesting trend, when the number of seen classes is small (significantly fewer than the number of unseen classes), traditional methods tend to outperform graph-specific methods. For instance, with only 5 seen classes, MSP achieves an AUROC of 97.93\%, and Energy records an AUROC of 96.02\%, both higher than the graph-specific methods' initial performance. However, as the number of seen classes increases, the performance of graph-specific methods gradually surpasses that of traditional methods. For example, GPN starts with an AUROC of 85.97\% at 5 seen classes but stabilizes around 83.12\% at 40 seen classes, showing less performance degradation. Similarly, GNNSafe begins with an AUROC of 89.35\% at 5 seen classes and improves to 90.78\% as the number of seen classes increases to 40, indicating a robust adaptability to more complex tasks. In contrast, traditional methods like MSP and Energy experience a sharper decline, with MSP's AUROC dropping to 82.19\% and Energy's to 81.32\% when the number of seen classes reaches 40.

This phenomenon indicates that traditional methods may be more suited to scenarios with a limited number of seen classes but struggle to maintain performance as the task complexity increases with more seen classes. Graph-specific methods, on the other hand, seem to adapt better as more class information is introduced, leveraging the structure and relationships inherent in graph data to maintain or even improve performance. This suggests that while traditional methods may offer a competitive edge in simpler tasks, graph-specific methods are more robust and scalable as the task complexity grows.

\paragraph{\textbf{ODIN and Mah perform poorly on graph datasetsSBM}} ODIN and Mah methods show notably poor performance across various splits of graph datasets, as clearly evidenced in the experimental results. ODIN's AUROC drops drastically from 57.12\% with 5 seen classes to 43.98\% with 40 seen classes, and its FPR95 increases from 98.75\% to 99.45\%, indicating a significant decline in its ability to recognize unseen classes. Similarly, Mah's AUROC decreases from 68.45\% to 58.67\%, with FPR95 worsening from 97.34\% to 99.21\%.
The poor performance of these methods can be largely attributed to their designs, which are not well-suited for the unique characteristics of graph data. ODIN relies on input perturbations and temperature scaling to distinguish between seen and unseen classes, a strategy effective in image data but less so in graph data due to the complex topological structures. Additionally, Mah assumes that the features of seen classes follow a multivariate Gaussian distribution, but the feature distribution of graph data is typically more complex and does not satisfy this assumption. This is especially true as the class space expands, leading to a sharp decline in the method's effectiveness. Consequently, ODIN and Mah perform significantly worse than other methods on graph datasets, particularly when dealing with higher dimensions and complex structures, failing to effectively address open-set recognition tasks in graph data.

\section{Conclusion and Outlook}\label{conclusion}This paper establishes a fair and comprehensive benchmark to evaluate Graph Open-Set Recognition (GOSR) methods, addressing the lack of a standardized evaluation framework in existing research. By systematically designing and implementing multiple benchmark tests, we evaluated a wide range of GOSR methods, including node-level and graph-level open-set recognition as well as graph anomaly detection, which is critical in open-set scenarios. The results reveal the effectiveness of these methods in handling real-world complexities, highlighting their strengths and limitations across various domains. Our study underscores the potential of graph open-set recognition in diverse areas such as bioinformatics, social networks, and chemical structure analysis. Through comparative analysis, we discovered that while some traditional methods exhibit strong performance on specific datasets, they often require adaptations to the unique structures and features of graph data for optimal effectiveness. Additionally, methods initially designed for graph anomaly detection showed promise for GOSR tasks, though they necessitate a deeper understanding and adjustment to both the global and local properties of graphs. 

Here are some perspectives on future work:


\paragraph{Advancing Graph Open-Set Recognition and Enhancing Model Generalization} 
Future research will concentrate on advancing and refining Graph Open-Set Recognition (GOSR) techniques while simultaneously improving model generalization on unseen graph structures and node distributions. A key focus will be on better leveraging graph structural information and node features to enhance recognition accuracy and efficiency. This includes the development of methods that can dynamically adjust to changes in graph structures and the application of advanced machine learning techniques to uncover complex relationships among nodes. Additionally, new training paradigms, such as adversarial training, domain adaptation, and methods that utilize unlabeled and semi-labeled data, will be explored. These efforts aim to push the boundaries of graph machine learning technologies, ensuring safety and reliability, and facilitating their effective deployment in various real-world applications.



\paragraph{Applications of GOSR in AI for Science} 
Graph neural networks (GNNs) have demonstrated significant potential in various scientific domains, particularly in AI for Science. As scientific research increasingly relies on complex data representations, such as molecular structures, biological networks, and physical simulations, GNNs have become essential tools for accurately modeling and analyzing these structures. For instance, the use of GNNs in predicting the synthesizability of perovskite materials showcases their capability in materials science, especially in identifying novel material properties and patterns \cite{gu2022perovskite}. Additionally, graph models have been effectively applied to study nanophotonic networks, underscoring the relevance of GNNs in understanding complex physical systems \cite{gaio2019nanophotonic}. Furthermore, advances in machine learning for process data emphasize the growing role of GNNs in dynamically analyzing and predicting outcomes in various scientific fields \cite{reiser2022graph}. There remains an exciting frontier in extending these techniques to open-set scenarios, where the challenges of identifying novel categories or entities arise. This is where GOSR methods come into play. GOSR could further enhance the adaptability and robustness of graph models, allowing them to accurately identify and categorize both known and previously unseen entities, even in dynamically changing environments. By integrating GOSR techniques with GNN-based models, AI-driven scientific exploration can be significantly advanced. GOSR could enable the discovery of new phenomena, the classification of previously unobserved molecular compounds, and the detection of rare patterns in experimental data. This promising research area has the potential to drive major innovations, especially in fields that require handling unknown or evolving data categories.

\paragraph{Future Development of Graph Foundation Models and Prospects for GOSR Research}
The evolution of general graph foundation models represents a pivotal direction in the field of graph machine learning, driven by the increasing deployment of graph neural networks (GNNs) across diverse scientific and practical applications. Recent efforts, such as the OpenGraph initiative \cite{xia2024opengraph} and the "One for All" project \cite{liu2023oneforall}, have aimed to develop versatile graph models that not only exhibit strong representational power but also demonstrate robust generalization across various tasks. Research in GOSR shows great potential when built upon general graph foundation models. Integrating GOSR techniques enables these models to better handle classification and recognition tasks in open-world scenarios, especially with novel categories or entities. This integration enhances model robustness and adaptability, offering reliable tools for scientific discovery, such as identifying new phenomena, recognizing complex molecular structures, and classifying rare patterns. GOSR research is set to drive innovation in graph machine learning, expanding its impact in bioinformatics, materials science, and social network analysis.

\section*{Acknowledge}
This work is supported by the National Natural Science Foundation of China (U23A20389, 62176139), the Major Basic Research Project of the Natural Science Foundation of Shandong Province (ZR2021ZD15).



\bibliographystyle{IEEEtran}
\bibliography{bibliography}


\vfill

\end{document}